\documentclass[nohyperref]{article}

\usepackage{microtype}
\usepackage{graphicx}
\usepackage{caption}
\usepackage{subcaption}
\usepackage{booktabs} 
\usepackage{multirow}

\usepackage{nicefrac}
\usepackage{floatrow}
\usepackage{appendix}
\usepackage{textpos}
\newfloatcommand{capbtabbox}{table}[][\FBwidth]

\usepackage[backref=page]{hyperref}

\usepackage{etoolbox}
\usepackage{xcolor}
\definecolor{mplgray}{HTML}{8a8a8a}
\renewcommand*{\backrefalt}[4]{%
    \ifcase #1 %
    No citations.%
    \or
    {\hypersetup{linkcolor=mplgray}{#2}}
    \else
    {\hypersetup{linkcolor=mplgray}{#2}}%
    \fi
}
\makeatletter
\let\oldBR@backref=\BR@backref
\renewcommand{\BR@backref}{\hfill\oldBR@backref}
\makeatother

\usepackage[dynnworkshop]{_includes/icml2022}
\usepackage{amsmath}
\usepackage{amssymb}
\usepackage{mathtools}
\usepackage{amsthm}

\usepackage[capitalize,noabbrev]{cleveref}
\theoremstyle{plain}

\theoremstyle{definition}

\theoremstyle{remark}

\newcommand{\dtoprule}{\specialrule{1pt}{0pt}{0.65pt}%
            \specialrule{0.3pt}{0pt}{\belowrulesep}%
            }
\newcommand{\dbottomrule}{\specialrule{0.3pt}{0pt}{0.65pt}%
\specialrule{1pt}{0pt}{\belowrulesep}%
}
\usepackage[textsize=tiny]{todonotes}

\begin{document}
\twocolumn [
\icmltitle{Sparse Relational Reasoning with Object-Centric Representations}




\begin{icmlauthorlist}
\icmlauthor{Alex F. Spies}{icl}
\icmlauthor{Alessandra Russo}{icl}
\icmlauthor{Murray Shanahan}{icl}
\end{icmlauthorlist}

\icmlaffiliation{icl}{Department of Computing, Imperial College London, London, United Kingdom}
\icmlcorrespondingauthor{Alex F. Spies}{afspies@imperial.ac.uk}

\icmlkeywords{Machine Learning, ICML, object-centric, computer-vision, reasoning}

\vskip 0.3in
] 



\printAffiliationsAndNotice{}  

\begin{abstract}
We investigate the composability of soft-rules learned by relational neural architectures when operating over object-centric (slot-based) representations, under a variety of sparsity-inducing constraints. We find that increasing sparsity, especially on features, improves the performance of some models and leads to simpler relations. Additionally, we observe that object-centric representations can be detrimental when not all objects are fully captured; a failure mode to which CNNs are less prone. These findings demonstrate the trade-offs between interpretability and performance, even for models designed to tackle relational tasks.
\end{abstract}

\section{Introduction}
Recent progress in deep learning has shown impressive results in many domains, ranging from reinforcement learning to natural language processing and image generation \cite{reedGeneralistAgent2022,brownLanguageModelsAre2020a,sahariaPhotorealisticTexttoImageDiffusion2022}. It has been argued that this reflects the importance of scale (in terms of dataset and model sizes) as opposed to sophisticated architectural design \cite{BitterLesson}; however, it remains unclear whether scaling up existing architectures will lead to generally capable artificial agents. In particular, many such approaches still struggle to systematically generalize, and to perform reasoning tasks that come easily to humans \cite{kocijanDefeatWinogradSchema2022,shanahanArtificialIntelligenceCommon2020}. These shortcomings have motivated much work in the object-centric and neuro-symbolic communities, which broadly speaking focus on ``inductively biasing" models toward learning or leveraging structured\footnote{As opposed to unstructured, i.e. monolithic, embeddings or network architectures.} representations \cite{greffBindingProblemArtificial2020a,garcezNeurosymbolicAI3rd2020a}.

Whilst such works are promising, it remains unclear to what extent \textit{structured reasoning over structured representations} occurs within some of the models proposed by the community. To this end, we investigate i) the representations and generalization behaviour of relational reasoning architectures when used in conjunction with object-centric encoders, and ii) whether encouraging sparsity at the level of learned relations (i.e. simpler relations) increases compositionality.


\section{Methodology}
There are two key aspects to our investigation. Firstly, we vary the structure of models' encoder outputs, using either slot-based representations or CNN feature maps. Secondly, we vary the kinds of sparsity priors present in a subset of the models; both at the level of feature-vector selection, and relations' feature dependencies. 

\subsection{Models}
\begin{figure*}
    \centering
    \includegraphics[width=0.95\textwidth]{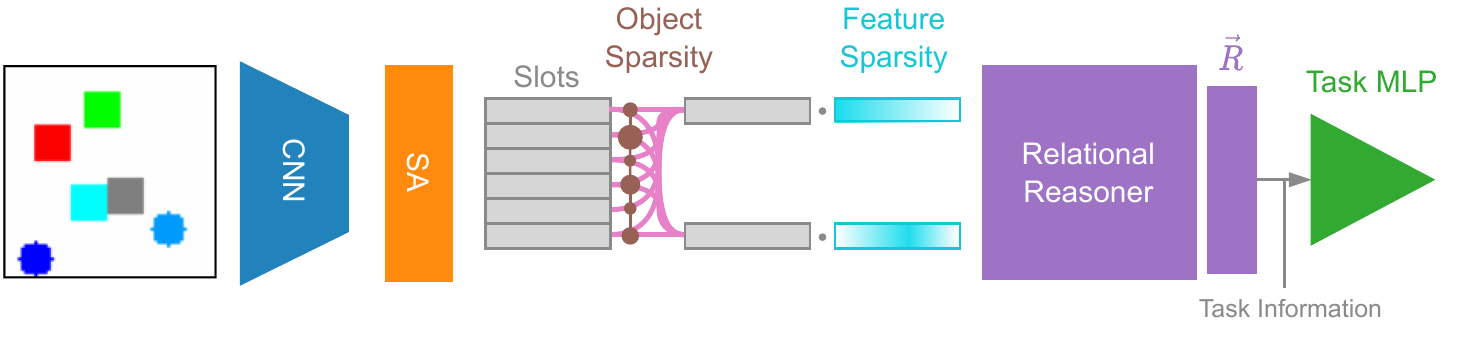}
    \vspace{-0.2cm}
    \caption{Schematic overview of a sparse relational reasoning architecture which leverages Slot Attention (SA). We impose two kinds of sparsity: i) Object sparsity is applied on the attention mechanism which selects objects (encoded in slots), and ii) Feature Sparsity is applied on the relations through regularization, encouraging sparse dependence on object-features for each learned relation.}
    \label{fig:method:arch}
\end{figure*}
An overview of the models investigated is shown in \cref{fig:method:arch}. For learning object-centric representations in the form of slots we use Slot Attention (SA) \cite{locatelloObjectCentricLearningSlot2020}. The outputs of SA or a simple CNN are then fed into one of two explicitly relational architectures: i) the RelationNet from \citet{santoroSimpleNeuralNetwork2017}, or ii) the PrediNet from \citet{shanahanExplicitlyRelationalNeural2019}. Task-relevant information (such as the question type) is then appended to the outputs of the ``relational reasoners'' before a small task-dependent MLP is applied. These architectures are chosen due to the relative simplicity with which they realise explicitly relational learning.

\textbf{The RelationNet} ensures that relational reasoning is performed by applying a shared MLP across all possible permutations of feature pairs and summing the outputs -  ensuring that the model is invariant to the order of the features.

\textbf{PrediNet:} Whereas the RelationNet considers all possible pairings of input features, the PrediNet selects ``two" (soft-attention is used) sets of features per ``head", and then learns a set of linear relations between these. As such, each head may be thought of as modelling a set of binary relations over feature vectors, with each element of a head's output vector, $\vec{R}$, corresponding to a single relation. The outputs of all heads are concatenated to form the output of the PrediNet.
\newpage
We modify the attention rule of the PrediNet in the case where slots are fed in - In this instance, queries are generated per slot, rather than using all features of all slots; we dub this ``Slot PrediNet".

\textbf{Slot Attention:} To extract object-centric representations from images we utilize Slot Attention (SA)~\cite{locatelloObjectCentricLearningSlot2020}. SA takes input features (e.g. from a CNN) of dimension $N\times D$, and maps these to $K$ output slots (vectors $\in \mathbb{R}^D$) . We pretrain the SA module through unsupervised reconstruction (autoencoding) , as in \citet{locatelloObjectCentricLearningSlot2020}. 

\subsection{Sparsity}
\label{sec:method:sparsity}
We experiment with two kinds of sparsity on the PrediNet. These are \textit{object sparsity} (relations specialize to few objects) and \textit{feature sparsity} (relations use few features per object).

\subsubsection{Inducing Sparsity}
\label{sec:method:sparsity:inducing}
\textbf{Object Sparsity} is applied at the level of the attention mechanism of the PrediNet, which is responsible for selecting the sets of features (or slots) over which each head applies relations. To encourage sparsity we replace the softmax attention with \texttt{sparsemax} \cite{martinsSoftmaxSparsemaxSparse2016} or \texttt{gumbel-softmax} \cite{jangCategoricalReparameterizationGumbelSoftmax2017a}. 

\textbf{Feature Sparsity} is applied at the level of the learned relation weights, which are applied to the (transformed) input features of each head. Pushing these weights to be sparse forces each head's relations to rely on as few elements in the input feature vectors (or slots) as possible. To this end, we apply L1 and Entropy regularization. 



\subsubsection{Measuring Sparsity}
\label{sec:method:sparsity:evaluating}
For each dimension of the output rule vector we fit a simple ancillary model which learns to map \textbf{two} input feature vectors to a scalar. This assumes that (i.e. confirms whether) the relational component can be well-approximated as a set of binary relations,
$\vec{R}^{(i)} = f(\vec{o}_j, \vec{o}_k)$, where the set of input feature-vectors is denoted as $\mathbf{O}=[\vec{o}_0,\dots ]$. 

We quantify the extent to which a relational module can be modelled in this fashion as the \text{deviation from binary relation} ($\Delta$BR).

\begin{figure}[h]
    \includegraphics[width=0.95\linewidth]{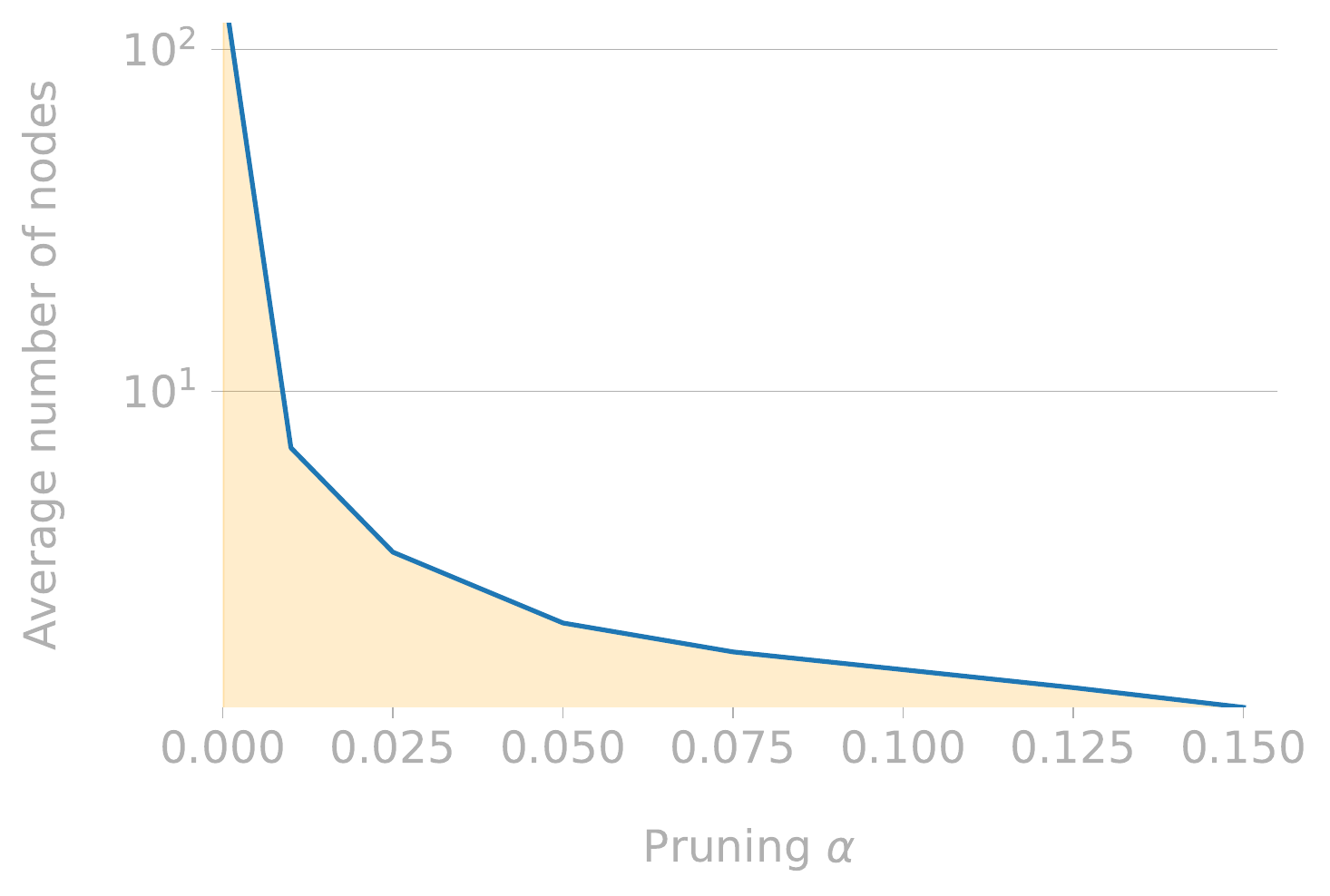}
    \vspace{-0.3cm}
    \caption{The number of nodes in the ancillary trees decreases as the degree of pruning (which trades off performance) is increased. The average number of nodes in the range $\alpha \in [0,0.15]$ is used as a measure of relation sparsity.}
    \label{fig:treenodes}
\end{figure}

The $\Delta$BR is calculated as the drop in performance of the end-to-end model when its relational reasoning module is replaced by a set of ancillary models. In particular, we chose to use Decision Trees owing to their simplicity and ease of pruning, which allow robustness to sparsity to be easily assessed, as shown in \cref{fig:treenodes}. Additionally, the dependence of a given decision tree on its inputs (``feature importances") can be used to quantify feature sparsity on the learned relations, by computing the entropy of these feature importances.

\section{Experiments}
\begin{table*}[h]
    \centering
    \caption{Test Accuracies (\%), across four seeds, for all model variants on both datasets (without sparsity priors). The Relations Game tasks are abbreviated as: Between Columns (BC), CP (Column Patterns), Row Patterns (RP) and Row Matching (RM). } 
    \label{tab:test_accuracies}
    \vskip 0.15in
    \begin{center}
    \begin{small}
    \begin{sc}
    \begin{tabular}{@{}lllllll@{}}
    \dtoprule
    \multicolumn{2}{c}{Model} & \multicolumn{3}{c}{Relations Game} & \multicolumn{2}{c}{Sort-of-CLEVR} \\\cmidrule(r){1-2}  \cmidrule(lr){3-5} \cmidrule(ll){6-7} 
    Encoder & Relational &  BC \textrightarrow CP &  BC \textrightarrow RM & RM \textrightarrow RP & Non-relational & Relational \\ \midrule
    SA      & MHA + MLP & $56 \pm 9$ & $50 \pm 2$ & $50 \pm 1$ & $85 \pm 3$ & $77 \pm 3$ \\
    CNN     & PrediNet & $87 \pm 3$ & $66 \pm 0$ & $74 \pm 8$ & $95 \pm 1$ & $79 \pm 0$ \\
    GT & Slot PrediNet & $86 \pm 2$ & $58 \pm 2$ & $85 \pm 2$ & $85 \pm 2$ & $81 \pm 1$ \\
    SA & Slot PrediNet & $73 \pm 2$ & $53 \pm 1$ & $59 \pm 3$ & $95 \pm 2$ & $77 \pm 2$ \\
    CNN & RelationNet & $53 \pm 1$ & $52 \pm 1$ & $52 \pm 1$ &  $100\pm0$  & $80\pm 1$ \\
    GT  & RelationNet & $75 \pm 12$ & $63 \pm 7$ & $55 \pm 8$ & $100 \pm 0$& $100 \pm 0$ \\
    SA  & RelationNet & $70 \pm 1$ & $49 \pm 1$ & $49 \pm 1$ & $100 \pm 0$ & $90 \pm 1$ \\ \dbottomrule
    \end{tabular}
    \end{sc}
    \end{small}
    \end{center}
    \vskip -0.1in
\end{table*}
Experiments are carried out on two visually simple datasets created to assess relational reasoning: i) the Relations Game from \citet{shanahanExplicitlyRelationalNeural2019}, and ii) Sort-of-CLEVR from \citet{santoroSimpleNeuralNetwork2017}. Curricula used for the Relations Game follow \citet{shanahanExplicitlyRelationalNeural2019}. See \cref{sec:app:datasets} for details.

\subsection{Do object-centric representations improve performance?}
We first investigate whether object-centric representations are beneficial when used in conjunction with relational reasoning architectures, as well as a simple multi-head attention baseline (MHA + MLP). To this end, we apply the PrediNet and RelationNet with i) learned CNN encoders, ii) ground-truth ``slots", and iii) pre-trained SA encoders.

The results of these experiments are summarized in \cref{tab:test_accuracies} and further discussion is provided in \cref{sec:app:results:discussion}. Focussing on the questions outlined above, we find that object-centric representations are not necessarily beneficial. Indeed, when used in conjunction with the PrediNet on the Relations Game, SA is detrimental. There are two likely reasons for this. Firstly, SA often fails to capture all objects present in a scene (see \cref{sec:app.slotatt.featurespace}), which is fatal for tasks in which knowledge of all objects is necessary; this explains why the Slot PrediNet does especially poorly on the RM\textrightarrow RP tasks, in which information about all six objects is required. Secondly, the CNN used with the PrediNet can learn relational representations directly, whereas SA encodes information about objects individually. 

An additional concern may be that SA fails to encode all relevant information about the objects that it does capture, but we show in \cref{sec:app.slotatt.assignment} that SA's representations are fairly complete. For the RelationNet we do observe a significant improvement in performance on relational tasks when using SA instead of a CNN. However, we attribute this primarily to the fact that the RelationNet struggles with tasks requiring simultaneous reasoning over more than two objects, and so is unable to learn useful CNN encodings.



\subsection{Does sparsity improve performance?}
\subsubsection{Assessing the degree of sparsity}
\label{sec:results:sparse.degree}
We first establish the degree to which the various sparsity biases discussed in \cref{sec:method:sparsity:inducing} lead to sparser relations.

\begin{figure*}[]
\includegraphics[width=0.85\textwidth]{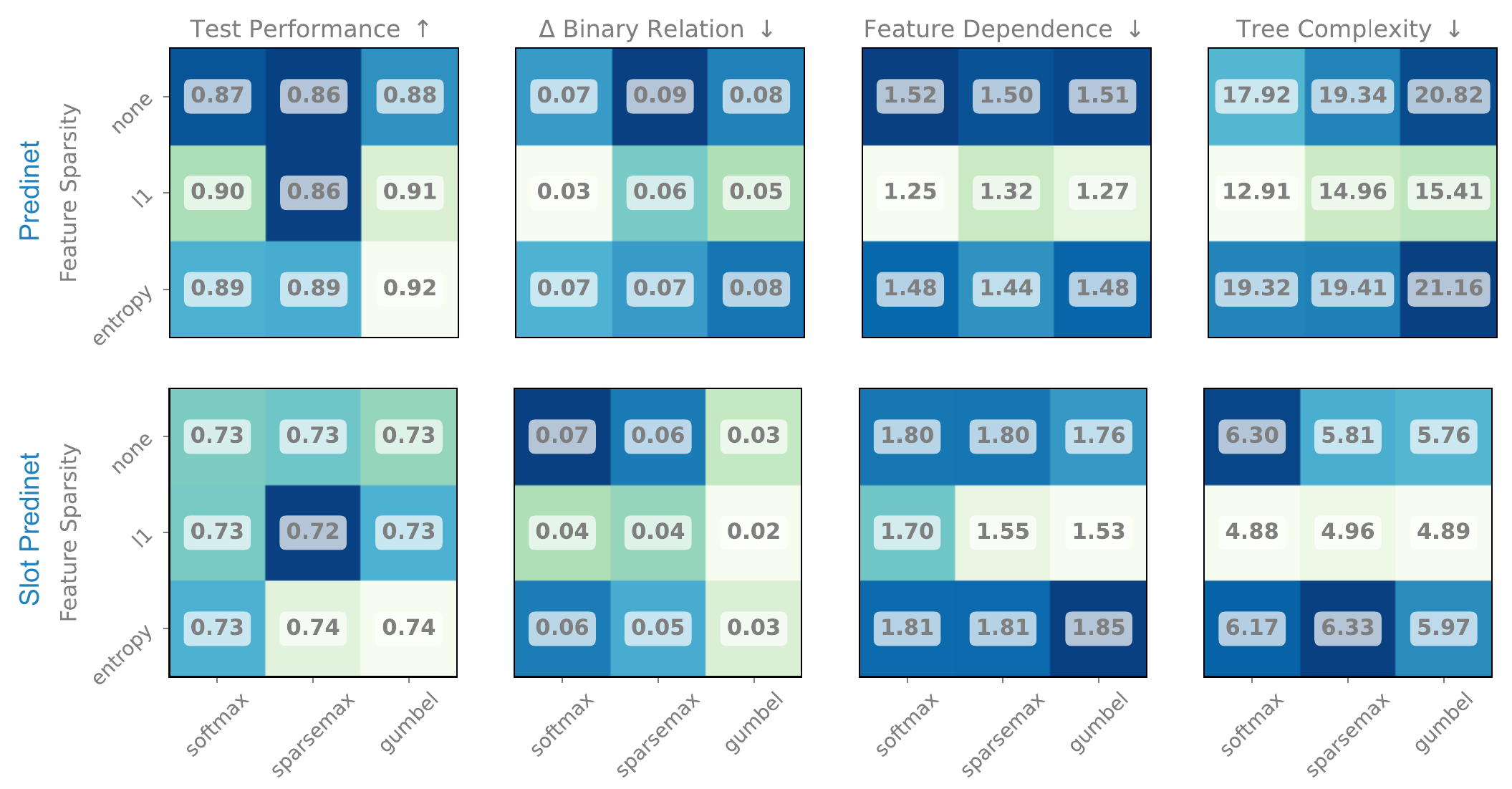}
\caption{We show metrics of interest for the PrediNet and Slot PrediNet for the Relations Game ``Between Columns" to ``Column Patterns" generalization task. All values correspond to averages on the generalization test set, taken over four seeds. The Feature Dependence and Tree Complexity metrics are averaged across all Decision Trees fitted to a given model's relational reasoning component.}
\label{fig:experiment:sparsity_plots}
\end{figure*}
In \cref{fig:experiment:sparsity_plots} we show the average feature dependence entropy for ancillary models (column three), as well as the average tree complexity across all levels of pruning (column four). For clarity, we show results only for the PrediNet with CNN and SA inputs, and only on the BC\textrightarrow CP Relations Game tasks. The complete suite of results is provided in \cref{sec:app:results:sparsity}, along with further discussion in \cref{sec:app:results:discussion}.

The key finding here is that regularizing the relations to rely on few features does indeed lead to simpler relations; in that, they may be faithfully modelled with Decision Trees consisting of fewer nodes, and relying upon fewer of the input features. To this end, L1 regularization appears considerably more effective than entropy regularization. 

This effect also holds for the Slot PrediNet, and it appears that the ancillary models are much simpler when slots are used. Whilst we might expect this result given that the slots' features are disentangled (see \cref{sec:app.slotatt.featurespace}), most of the effect is likely due to the task performance difference between the models; that is to say, as the Slot PrediNet performs worse, its relations may be modelled more simply.


\subsubsection{Assessing the impact of sparsity}
\label{sec:results:sparse.impact}
Having established that the sparsity priors do in fact induce sparsity, at least when it comes to features, we investigate whether their effects are desirable.
To this end, we are interested in the first two columns of \cref{fig:experiment:sparsity_plots} which show the performance on the test set for models whose relational modules have been i) left unchanged (i.e. remain neural) and ii) replaced by a collection of Decision Trees (shown as the difference from column 1).

Again there are two primary findings. Firstly, increasing sparsity improves the performance of the models on the RelationNet generalization tasks, however, this effect is more pronounced for the PrediNet than the Slot PrediNet. This is not surprising, as the PrediNet can easily mix information from all features of all objects, and thus benefits most from being biased towards increased sparsity. 

Secondly, looking at the $\Delta$BR we see that L1 regularization leads to relations more closely approximating independent binary rules. However, the different forms of object sparsity have little systematic impact. We primarily attribute this to the similarity of soft and hard-attention values by the end of training. It is worth noting that hard attention appears to be beneficial when used in conjunction with the PrediNet; likely as this prevents the CNN from learning relational features which overfit to the pretraining task. 

\section{Related Work}
There are other relational reasoning architectures besides those tested here. Most notably, Neural Production Systems \cite{goyalNeuralProductionSystems2021a} which operate over slot-based representations and enforce object and relation-application sparsity. Additionally, graph neural networks are also (dense) relational reasoners, and have been applied in conjunction with object-centric encoders~\cite{kipfContrastiveLearningStructured2020}.




\section{Conclusions}
In this work we investigated the extent to which sparse representations pair well with relational reasoning modules. We consider our primary findings to be that \textbf{object-centric representations are not necessarily beneficial} and that \textbf{relational reasoning modules are highly sensitive} to their input representations, \textit{and} the forms of relational tasks required of them. 

This suggests that future work in this area should be wary of overstating model capabilities before testing on a diverse range of tasks.

In particular, we demonstrated that whilst perfect object-centric representations are beneficial, imperfect representations may be disproportionately detrimental. However, encouraging sparsity, especially on the features utilized by learned relations, is beneficial to both task performance and relation simplicity. Sparsity applied at the level of objects is less effective, though it remains unclear to what extent this is a symptom of imperfect object representations, or simply a demonstration of soft-attention's tendency to approximate hard-attention as training progresses. Additional, whilst increasing sparsity does improve model simplicity, the resulting relations are still quite complex\footnote{In the sense that they are only well-approximated by Decision Trees with many nodes, as per the final column of \cref{fig:experiment:sparsity_plots}.} and thus limited in their interpretability. 



\textbf{Future work} may aim to provide robustness to incomplete object-centric representations by implementing a form of residual connection bypassing the object-centric learner. Additionally, it is worth exploring the use of weak supervision to guide learned relations to utilize all potentially informative parts of a disentangled object-centric representation. Lastly, we did not investigate relation-application sparsity\footnote{This could be realised by applying a learned mask, or additional attention mechanism, after the relational learning module.}, which may further encourage compositionality.




\clearpage
\bibliographystyle{_includes/icml2022}
\bibliography{refs.bib}

\begin{thebibliography}{25}
\providecommand{\natexlab}[1]{#1}
\providecommand{\url}[1]{\texttt{#1}}
\expandafter\ifx\csname urlstyle\endcsname\relax
  \providecommand{\doi}[1]{doi: #1}\else
  \providecommand{\doi}{doi: \begingroup \urlstyle{rm}\Url}\fi

\bibitem[Abadi et~al.(2015)Abadi, Agarwal, Barham,
  et~al.]{tensorflow2015-whitepaper}
Abadi, M., Agarwal, A., Barham, P., et~al.
\newblock {TensorFlow}: Large-scale machine learning on heterogeneous systems,
  2015.
\newblock URL \url{https://www.tensorflow.org/}.
\newblock Software available from tensorflow.org.

\bibitem[Bradbury et~al.(2018)Bradbury, Frostig, Hawkins,
  et~al.]{jax2018github}
Bradbury, J., Frostig, R., Hawkins, P., et~al.
\newblock {JAX}: composable transformations of {P}ython+{N}um{P}y programs,
  2018.
\newblock URL \url{http://github.com/google/jax}.

\bibitem[Breiman et~al.(1984)Breiman, Friedman, Olshen, and
  Stone]{breimanClassificationRegressionTrees2017}
Breiman, L., Friedman, J.~H., Olshen, R.~A., and Stone, C.~J.
\newblock \emph{Classification {{And Regression Trees}}}.
\newblock {Routledge}, {New York}, October 1984.
\newblock ISBN 978-1-315-13947-0.

\bibitem[Brown et~al.(2020)Brown, Mann, Ryder,
  et~al.]{brownLanguageModelsAre2020a}
Brown, T.~B., Mann, B., Ryder, N., et~al.
\newblock Language {{Models}} are {{Few-Shot Learners}}, 2020.
\newblock URL \url{http://arxiv.org/abs/2005.14165}.

\bibitem[d'Avila Garcez \& Lamb(2020)d'Avila Garcez and
  Lamb]{garcezNeurosymbolicAI3rd2020a}
d'Avila Garcez, A. and Lamb, L.~C.
\newblock Neurosymbolic {{AI}}: {{The}} 3rd {{Wave}}.
\newblock \emph{{arXiv}}, December 2020.
\newblock URL \url{http://arxiv.org/abs/2012.05876}.

\bibitem[Eastwood \& Williams(2018)Eastwood and
  Williams]{eastwoodFrameworkQuantitativeEvaluation2018a}
Eastwood, C. and Williams, C. K.~I.
\newblock A {{Framework}} for the {{Quantitative Evaluation}} of {{Disentangled
  Representations}}.
\newblock \emph{{{ICLR}}}, 2018.
\newblock URL \url{https://openreview.net/forum?id=By-7dz-AZ}.

\bibitem[Goyal et~al.(2021)Goyal, Didolkar, Ke,
  et~al.]{goyalNeuralProductionSystems2021a}
Goyal, A., Didolkar, A., Ke, N., et~al.
\newblock Neural {{Production Systems}}.
\newblock In \emph{NeurIPS}, 2021.
\newblock URL \url{https://openreview.net/forum?id=xQGYquca0gB}.

\bibitem[Greff et~al.(2020)Greff, {van Steenkiste}, and
  Schmidhuber]{greffBindingProblemArtificial2020a}
Greff, K., {van Steenkiste}, S., and Schmidhuber, J.
\newblock On the {{Binding Problem}} in {{Artificial Neural Networks}},
  December 2020.
\newblock URL \url{http://arxiv.org/abs/2012.05208}.

\bibitem[Hennigan et~al.(2020)Hennigan, Cai, Norman, and
  Babuschkin]{haiku2020github}
Hennigan, T., Cai, T., Norman, T., and Babuschkin, I.
\newblock {H}aiku: {S}onnet for {JAX}, 2020.
\newblock URL \url{http://github.com/deepmind/dm-haiku}.

\bibitem[Hessel et~al.(2020)Hessel, Budden, Viola, Rosca, Sezener, and
  Hennigan]{optax2020github}
Hessel, M., Budden, D., Viola, F., Rosca, M., Sezener, E., and Hennigan, T.
\newblock Optax: composable gradient transformation and optimisation, in jax!,
  2020.
\newblock URL \url{http://github.com/deepmind/optax}.

\bibitem[Jang et~al.(2017)Jang, Gu, and
  Poole]{jangCategoricalReparameterizationGumbelSoftmax2017a}
Jang, E., Gu, S., and Poole, B.
\newblock Categorical {{Reparameterization}} with {{Gumbel-Softmax}}.
\newblock \emph{ICLR}, 2017.
\newblock URL \url{http://arxiv.org/abs/1611.01144}.

\bibitem[Johnson et~al.(2017)Johnson, Hariharan, Van Der~Maaten, et~al.]{clevr}
Johnson, J., Hariharan, B., Van Der~Maaten, L., et~al.
\newblock {CLEVR: A diagnostic dataset for compositional language and
  elementary visual reasoning}.
\newblock \emph{IEEE conference on CV and PR}, 2017.
\newblock URL \url{https://arxiv.org/abs/1612.06890}.

\bibitem[Kipf et~al.(2020)Kipf, {van der Pol}, and
  Welling]{kipfContrastiveLearningStructured2020}
Kipf, T., {van der Pol}, E., and Welling, M.
\newblock Contrastive {{Learning}} of {{Structured World Models}}.
\newblock \emph{{ICLR}}, 2020.
\newblock URL \url{http://arxiv.org/abs/1911.12247}.

\bibitem[Kocijan et~al.(2022)Kocijan, Davis, Lukasiewicz,
  et~al.]{kocijanDefeatWinogradSchema2022}
Kocijan, V., Davis, E., Lukasiewicz, T., et~al.
\newblock The {{Defeat}} of the {{Winograd Schema Challenge}}, 2022.
\newblock URL \url{http://arxiv.org/abs/2201.02387}.

\bibitem[Locatello et~al.(2020)Locatello, Weissenborn, Unterthiner,
  et~al.]{locatelloObjectCentricLearningSlot2020}
Locatello, F., Weissenborn, D., Unterthiner, T., et~al.
\newblock Object-{{Centric Learning}} with {{Slot Attention}}.
\newblock 2020.
\newblock URL \url{http://arxiv.org/abs/2006.15055}.

\bibitem[Martins \& Astudillo(2016)Martins and
  Astudillo]{martinsSoftmaxSparsemaxSparse2016}
Martins, A. F.~T. and Astudillo, R.~F.
\newblock From {{Softmax}} to {{Sparsemax}}: {{A Sparse Model}} of
  {{Attention}} and {{Multi-Label Classification}}.
\newblock \emph{ICML}, 2016.
\newblock URL \url{http://arxiv.org/abs/1602.02068}.

\bibitem[Pedregosa et~al.(2011)Pedregosa, Varoquaux, Gramfort,
  et~al.]{scikit-learn}
Pedregosa, F., Varoquaux, G., Gramfort, A., et~al.
\newblock Scikit-learn: Machine learning in {P}ython, 2011.

\bibitem[Reed et~al.(2022)Reed, Zolna, Parisotto,
  et~al.]{reedGeneralistAgent2022}
Reed, S., Zolna, K., Parisotto, E., et~al.
\newblock A {{Generalist Agent}}.
\newblock \emph{{arXiv}}, May 2022.
\newblock URL \url{http://arxiv.org/abs/2205.06175}.

\bibitem[Saharia et~al.(2022)Saharia, Chan, Saxena,
  et~al.]{sahariaPhotorealisticTexttoImageDiffusion2022}
Saharia, C., Chan, W., Saxena, S., et~al.
\newblock Photorealistic {{Text-to-Image Diffusion Models}} with {{Deep
  Language Understanding}}.
\newblock \emph{{arXiv}}, May 2022.
\newblock URL \url{http://arxiv.org/abs/2205.11487}.

\bibitem[Santoro et~al.(2017)Santoro, Raposo, Barrett,
  et~al.]{santoroSimpleNeuralNetwork2017}
Santoro, A., Raposo, D., Barrett, D. G.~T., et~al.
\newblock A simple neural network module for relational reasoning.
\newblock \emph{NIPS}, 2017.
\newblock URL \url{http://arxiv.org/abs/1706.01427}.

\bibitem[Shanahan et~al.(2020{\natexlab{a}})Shanahan, Crosby, Beyret, and
  Cheke]{shanahanArtificialIntelligenceCommon2020}
Shanahan, M., Crosby, M., Beyret, B., and Cheke, L.
\newblock Artificial {{Intelligence}} and the {{Common Sense}} of {{Animals}}.
\newblock \emph{Trends in Cog.Sci.}, 24\penalty0 (11):\penalty0 862--872,
  November 2020{\natexlab{a}}.
\newblock URL
  \url{https://www.cell.com/trends/cognitive-sciences/abstract/S1364-6613(20)30216-3}.

\bibitem[Shanahan et~al.(2020{\natexlab{b}})Shanahan, Nikiforou, Creswell,
  et~al.]{shanahanExplicitlyRelationalNeural2019}
Shanahan, M., Nikiforou, K., Creswell, A., et~al.
\newblock An {{Explicitly Relational Neural Network Architecture}}.
\newblock \emph{ICML}, 2020{\natexlab{b}}.
\newblock URL \url{http://arxiv.org/abs/1905.10307}.

\bibitem[Sutton(2019)]{BitterLesson}
Sutton, R.
\newblock The {{Bitter Lesson}}, March 2019.
\newblock URL \url{http://www.incompleteideas.net/IncIdeas/BitterLesson.html}.

\bibitem[Tripathi(2020)]{socdataset}
Tripathi, R.~M.
\newblock Sort-of-clevr reimplementation.
\newblock \url{https://github.com/RishikMani/Sort-of-CLEVR}, 2020.

\bibitem[Watters et~al.(2019)Watters, Matthey, Burgess, and
  Lerchner]{wattersSpatialBroadcastDecoder2019}
Watters, N., Matthey, L., Burgess, C.~P., and Lerchner, A.
\newblock Spatial {{Broadcast Decoder}}: {{A Simple Architecture}} for
  {{Learning Disentangled Representations}} in {{VAEs}}.
\newblock \emph{ICLR LLD Workshop}, 2019.
\newblock URL \url{http://arxiv.org/abs/1901.07017}.

\end{thebibliography}

\newpage
\onecolumn
\appendix
\begin{figure}[!t]
\begin{minipage}{\textwidth}
\icmltitle{\vspace{-0.1cm}Appendix\\\vspace{0.15cm}\textnormal{Sparse Relational Reasoning with Object-centric Representations\vspace{-0.1cm}}}
\end{minipage}
\end{figure}

\section{Slot-Attention Encoder}
\subsection{Object property encoding in slots}
\label{sec:app.slotatt.assignment}
We investigate the extent to which ground truth factors of variations are well captured by the Slot Attention latents. This is challenging, as when utilizing slot-based representations, multiple latent vectors encode information about the image, and it is not always known which latent corresponds to which feature. To address this we use the model's attention maps to perform slot-object pairing. In particular, we use openCV to ﬁnd contours in the normalized attention masks, and then apply an adaptive threshold to establish whether there is: i) exactly one contour in the map, and ii) whether the contour area is less than a ﬁxed fraction of the total image (a heuristic for discarding background slots). The center position of the resulting contour is then used to match the slot to a ground truth object, the properties of which are assigned as labels to the corresponding slot. \cref{fig:app.slotatt.size.assign} shows examples of this assignment procedure.

After this labelling procedure has been performed, we collect 20,000 slots with associated object labels. Following \citet{wattersSpatialBroadcastDecoder2019}, we then train ancillary models (Gradient Boosted Regressors) to predict ground truth factors of variation, such as position, from these slots. By querying the feature importances of the resulting models, and assessing their prediction accuracies on held-out slots, we are able to assess the representation of objects within slots.

\begin{figure}[h]
    \includegraphics[width=0.8\textwidth]{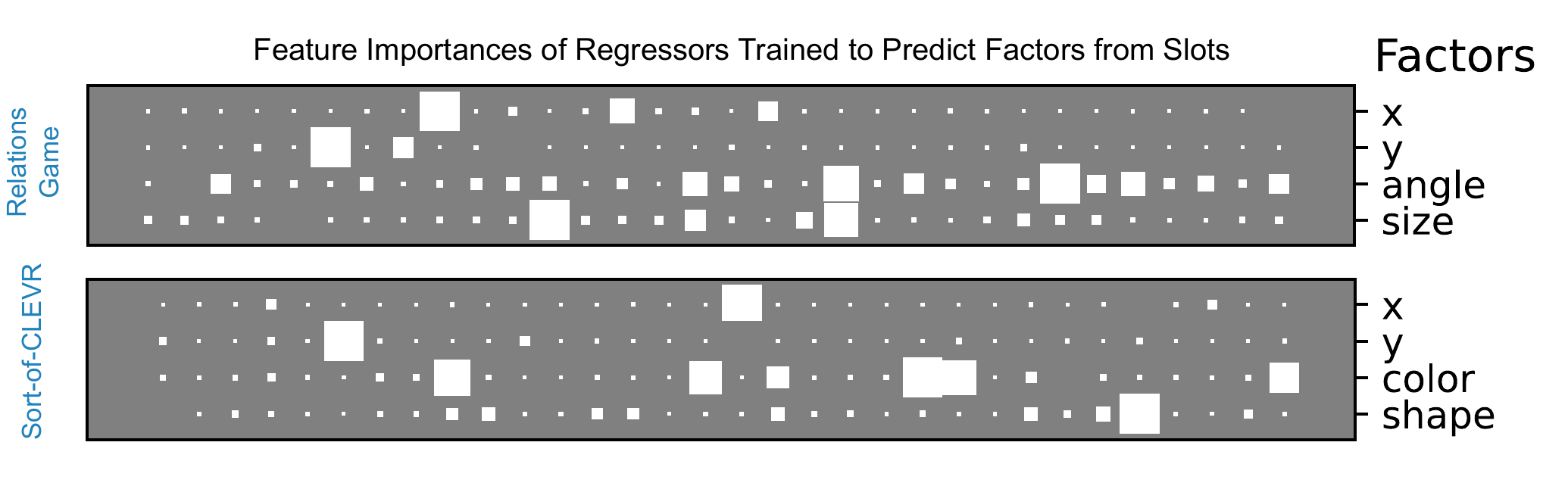}
    \caption{Hinton diagram showing disentanglement properties of the Slot Attention latents. Completeness corresponds to the average one-hotness of the columns, and likewise along rows for disentanglement. Broadly speaking, position information is strongly disentangled, as expected when using a spatial-broadcast decoder to pre-training Slot Attention \cite{wattersSpatialBroadcastDecoder2019}.}
    \label{fig:app:slot_att_feats}
\end{figure}

The feature importances for multiple generative factors are then combined into a matrix, as shown in \cref{fig:app:slot_att_feats}. Using this matrix, we compute the completeness and disentanglement of the slot representations, shown in \cref{tab:latent_fits}. These are defined as:
\textbf{Completeness} - the extent to which a single latent encodes information about only a single factor of variation, and \textbf{Disentanglement} - the extent to which all information about a single factor of variation is captured in only one latent.

\begin{table}[h!]
    \centering
    \caption{Latent properties obtained using the slot-object matching procedure outlined in \cref{sec:app.slotatt.featurespace}. The completeness and disentanglement metrics are computed following the definitions of \citet{eastwoodFrameworkQuantitativeEvaluation2018a}, and the Gradient-boosted Regressors were used as ancillary models. Random performance on Size prediction on Relations Game would equal 33\%. }
    \label{tab:latent_fits}
    \vskip 0.15in
    \begin{center}
    \begin{small}
    \begin{sc}
    \resizebox{\textwidth}{!}{%
    \begin{tabular}{lllllllll}
    \dtoprule
    &  & \multicolumn{2}{c}{Latent Structure Measures}      & \multicolumn{5}{c}{Ancillary Model Accuracy}  \\ \cmidrule(lr){3-4} \cmidrule(l){5-9}
    \multicolumn{1}{c}{Model} & \multicolumn{1}{c}{Dataset} & Completeness     & Disentanglement  & x  & y    & shape  & color  & size \\ \cmidrule(){1-9}
    \multirow{2}{*}{Small SA}    & Relations Game            & 0.33          & 0.55                     & 77 & 75   & -      & -      & 51            \\
                                 & Sort-of-CLEVR             & 0.40          & 0.62                     & 93 & 92   & 90     & 92     & - \vspace{0.2cm} \\
                                 
    \multirow{2}{*}{Big SA} & Relations Game                & 0.43          & 0.62                     & 84 & 86   & -      & -      & 69            \\
                            & Sort-of-CLEVR                  & 0.49          & 0.82                     & 95 & 95   & 96     & 91     & -             \\
    \dbottomrule           
    \end{tabular}
    }
    \end{sc}
    \end{small}
    \end{center}
    \vskip -0.1in
\end{table}

\begin{figure}[h!]
\includegraphics[width=0.85\textwidth]{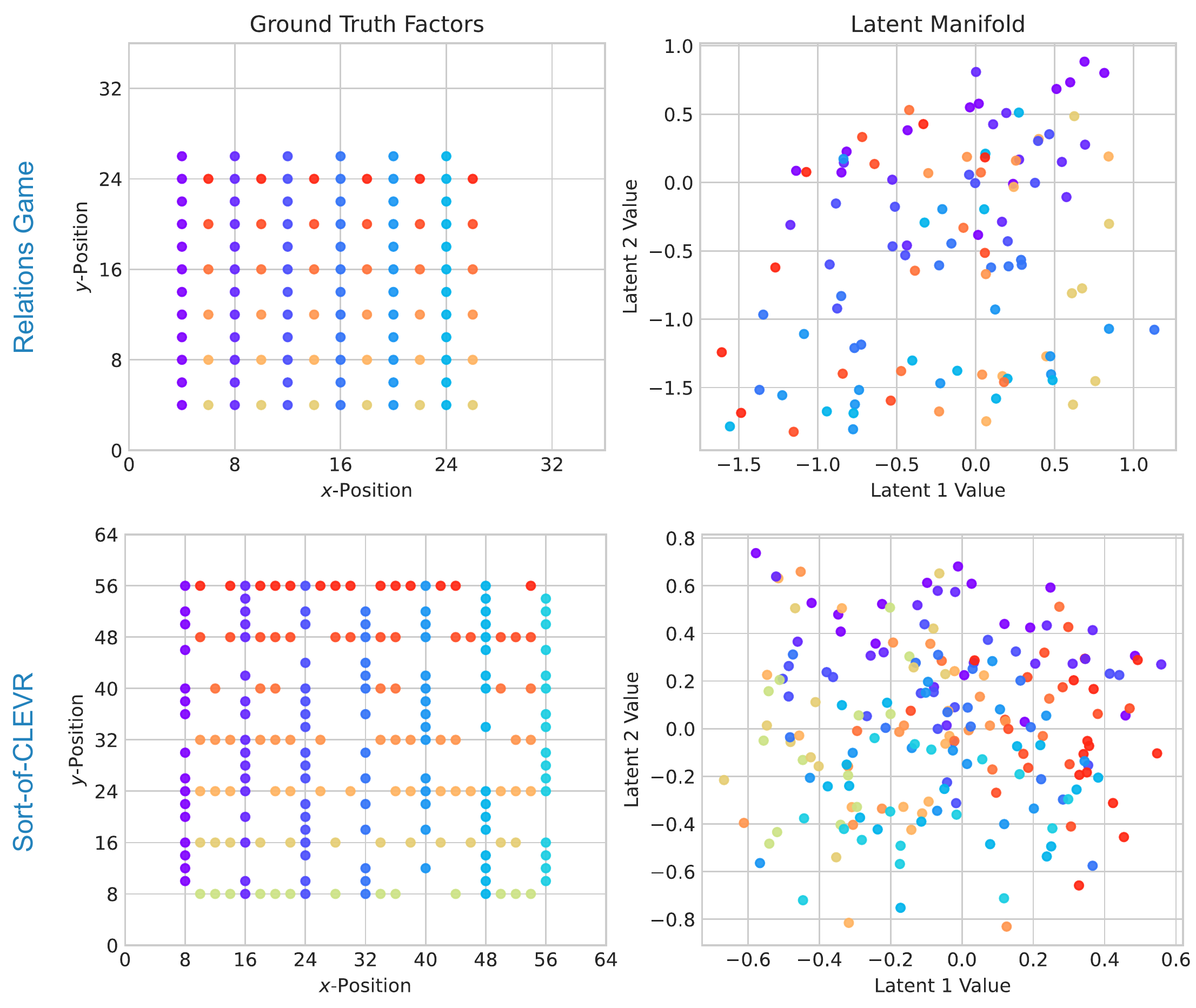}
\caption{For the small SA model, the dimensions deemed to most saliently encode x and y position are compared to their ground-truth equivalents. This shows that SA learns an approximately linear mapping for position information (up to a rotation). The apparent noisyness of the latent manifolds is primarily due to the imperfectness of the slot-object matching procedure. Missing points correspond to cases where no matching slot was found (from the subset of the test-set considered).}
\label{fig:app:slot_attn_manifolds}
\end{figure}



\subsection{Comparison between low and high dimensional Slot-Attention}
\label{sec:app.slotatt.featurespace}
\begin{figure}[h]
\includegraphics[width=\textwidth]{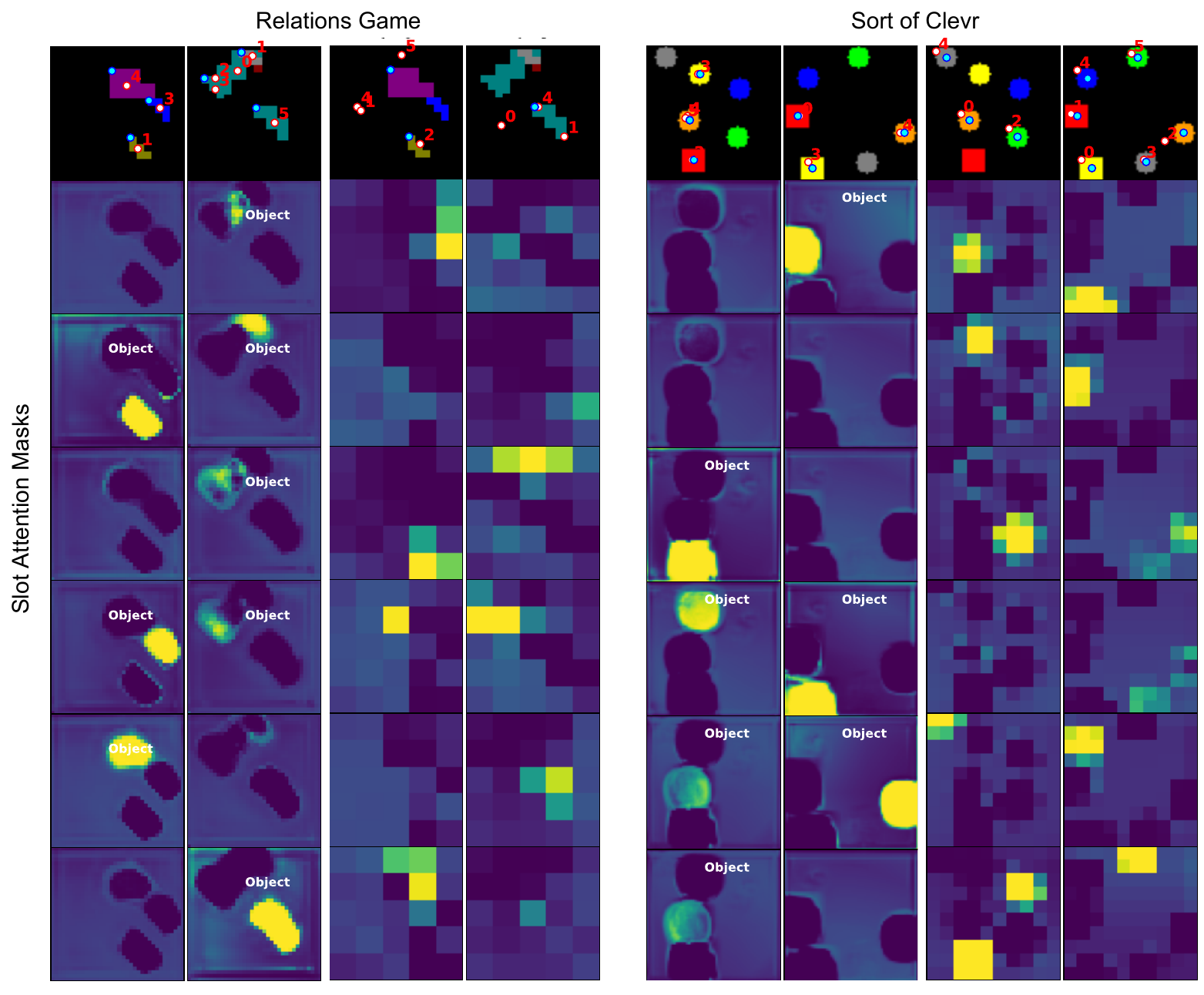}
\caption{Large and small Slot Attention models are compared. The top row shows the input images, and the remaining rows show the attention masks of the SA module. The first two columns for each dataset correspond to the full-size (latent resolution matching image resolution) SA models. Additionally, the slot-object matching procedure described above is shown. For the large model, masks in which contiguous contours have been detected are labelled with ``object", and the numbers and positions (red) on the images show the centroids of the mask contours with corresponding indices (0 for top row, and so on). The blue points denote ground-truth object positions.}
\label{fig:app.slotatt.size.assign}
\end{figure}

To fairly compare the slot-attention based models against ones with a CNN encoder, it was necessary to train a version of the model with a latent feature space of lower resolution than the input images. \cref{fig:app.slotatt.size.assign} shows the qualitative difference in model behaviour, and \cref{tab:latent_fits} shows differences in measured latent structure. Essentially, both models capture relevant information, and differences in quantitative measures likely reflect shortcomings of the slot-object assignment procedure.

It is worth noting that the inaccuracy of the assignment procedure can be seen both in the manifolds in \cref{fig:app:slot_attn_manifolds} as well as the mask fitting examples in \cref{fig:app.slotatt.size.assign}. This inaccuracy is more dramatic for the small SA model on the relations game, as contours may be very crude, and thus nearby objects may be matched to incorrect slots. We suspect this accounts for the majority of the difference in \textit{measured disentanglement} between the two models' latent spaces.

\clearpage
\begin{figure}[h]
    \includegraphics[width=0.5\linewidth]{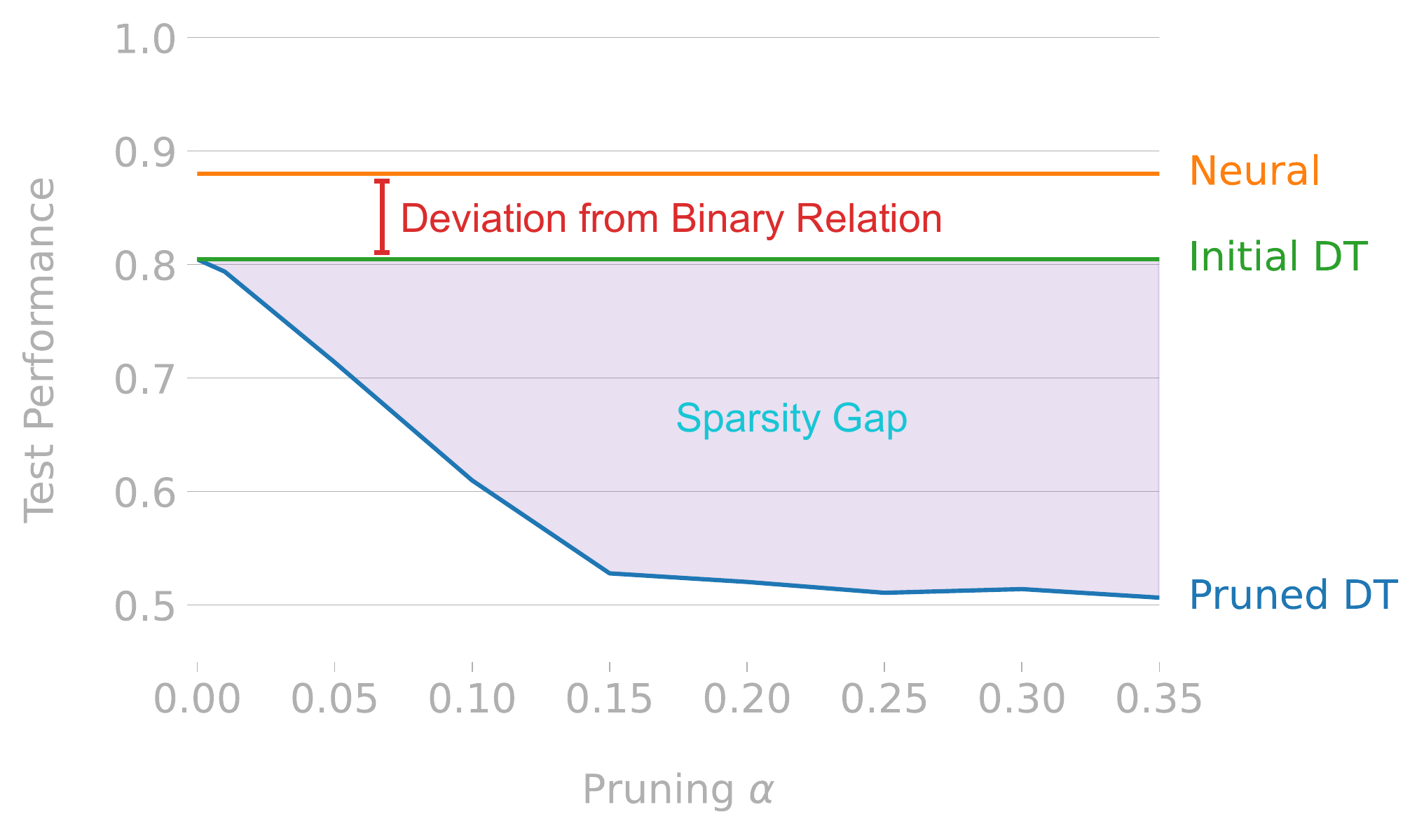}
    \label{fig:sparsity_metric}
    \caption{Visualization of sparsity measures relying upon the sweep of Decision Tree (DT) pruning complexities.}
\end{figure}

\section{Additional Results}
\label{sec:app:results:sparsity}
\begin{figure}[h!]
    \includegraphics[width=.95\linewidth]{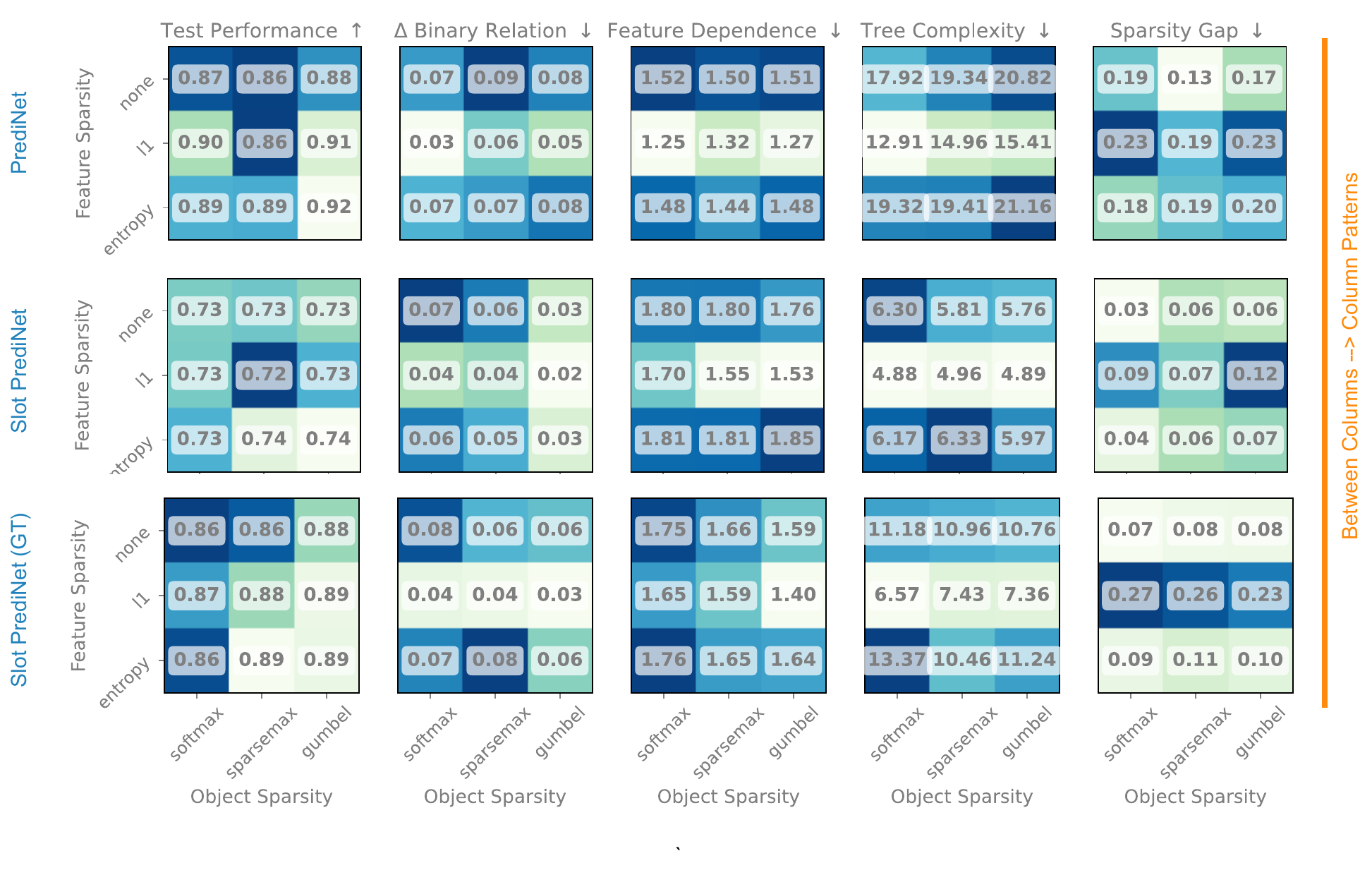}
    \vspace{-0.3cm}
    \caption{Sparsity sweeps on the Between Columns  \textrightarrow Column Patterns curriculum.}
    \label{fig:app:sparsity:betweencol}
\end{figure}

\begin{figure}[h]
    \includegraphics[width=.95\linewidth]{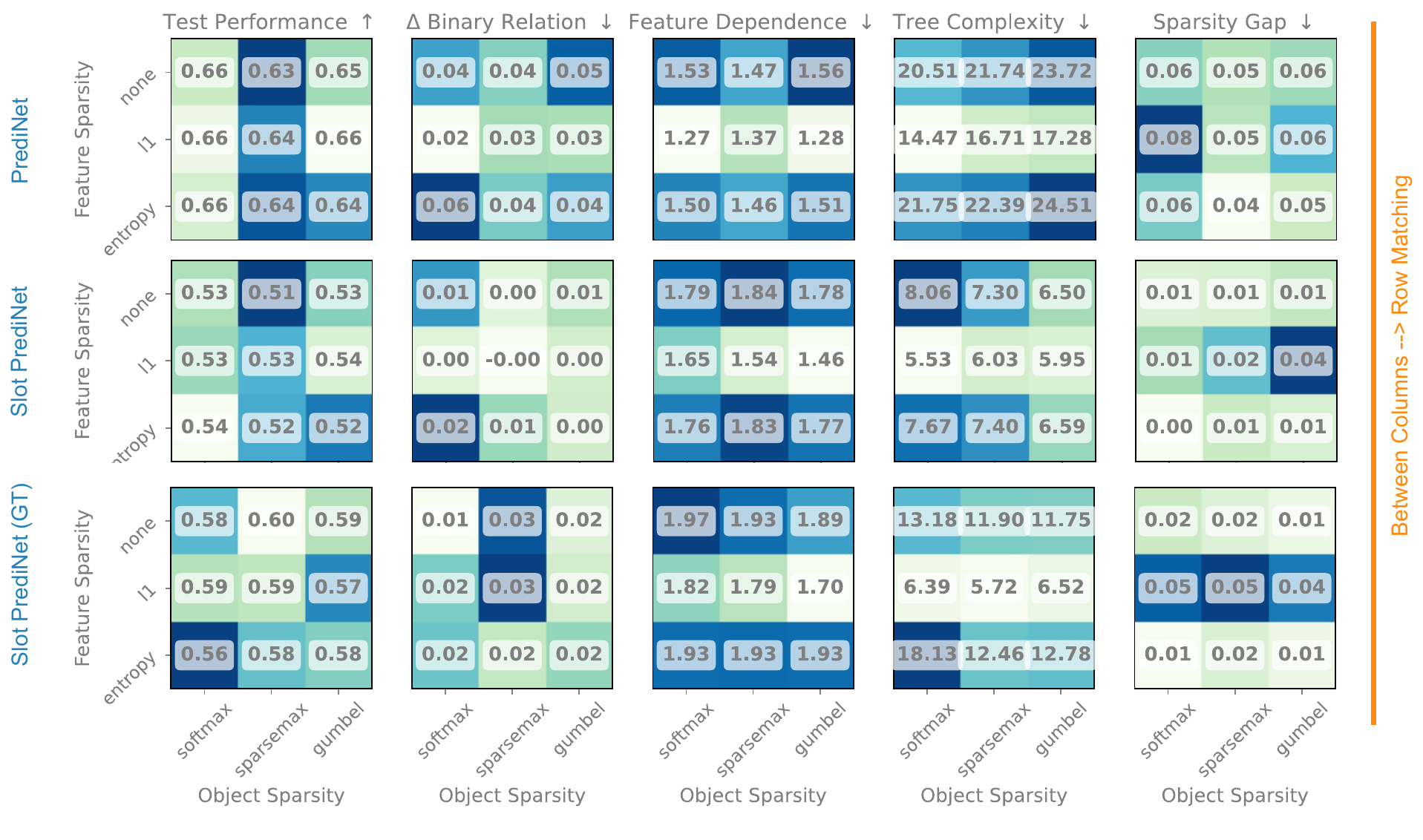}
    \caption{Sparsity sweeps on the Between Columns \textrightarrow Row Matching curriculum.}
    \label{fig:app:sparsity:rowmatch}
\end{figure}

\begin{figure}[h]
    \includegraphics[width=.95\linewidth]{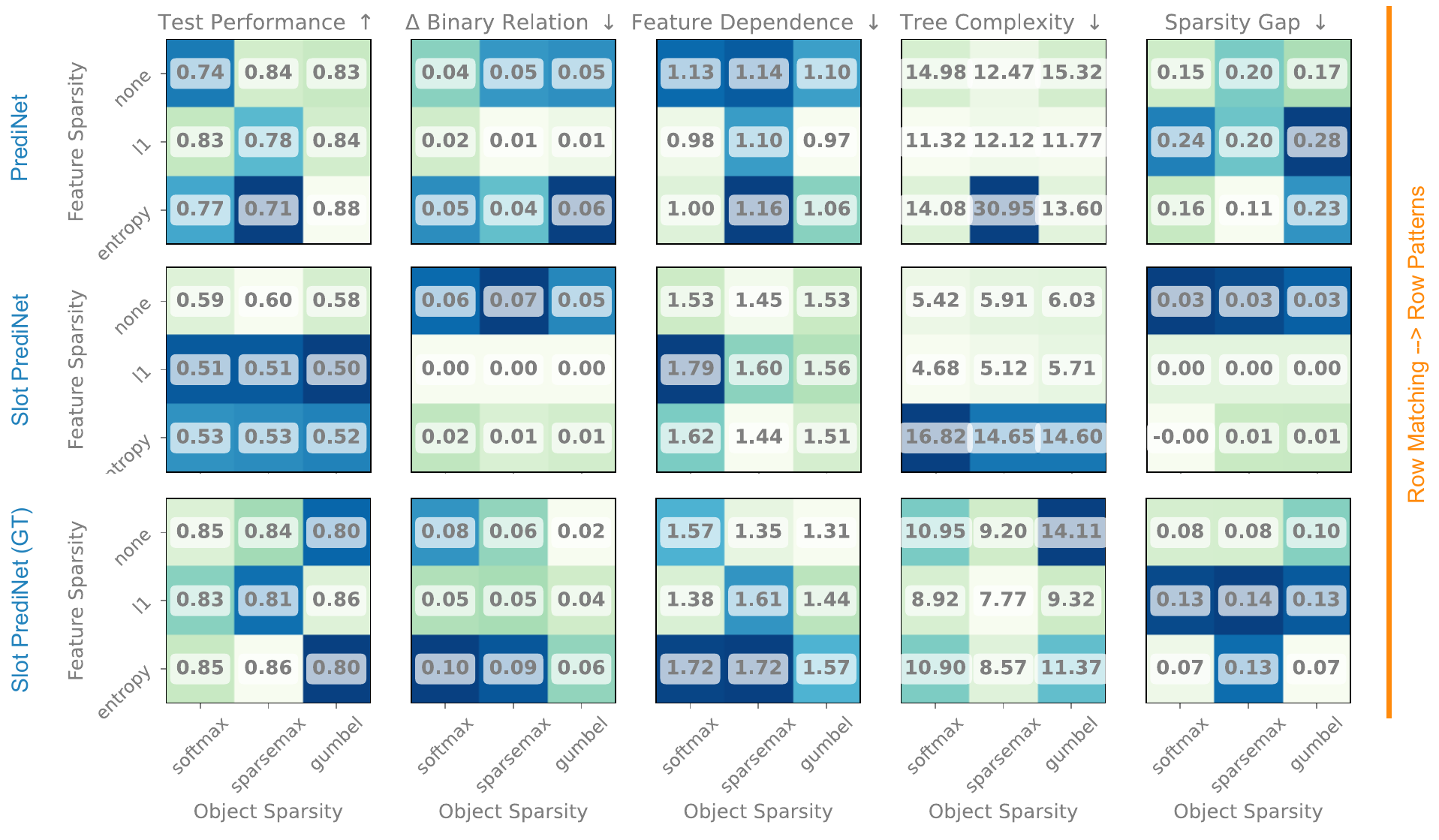}
    \caption{Sparsity sweeps on the  Row Matching \textrightarrow Row Patterns curriculum.}
    \label{fig:app:sparsity:rowpattern}
\end{figure}

Here we provide complete results for the sparsity sweeps carried out on the relations game curricula, as shown in \cref{fig:app:sparsity:betweencol,fig:app:sparsity:rowmatch,fig:app:sparsity:rowpattern}. We also report the ``Sparsity Gap" - defined as the average drop in performance across pruning (see \cref{fig:sparsity_metric}).


\clearpage
\subsection{Additional Discussion}
\label{sec:app:results:discussion}
Whilst we have seen that there are few general conclusions which drawn be made about the properties of the relational reasoning architectures we tested, there are points to be made which apply to the PrediNet, Slot PrediNet and RelationNet individually.

First we note that the sparsity gap metric mostly correlates with the test performance - a model which performs very poorly prior to pruning or ancillary model replacement will show little drop in performance during pruning. As such, we omitted discussion of the sparsity gap in the main paper. In fact, the same is true to some extent of all sparsity metrics considered. As such, we focussed our discussion on the BC\textrightarrow CP curriculum, in which all models performed well.

\textbf{Limitations of one-step relation application:} Both the PrediNet and RelationNet are unable to re-apply the same rule multiple times (i.e. recursion), and so cannot straightforwardly handle cases where the same relationship is required many times. The PrediNet is able to re-learn the same relationship across heads, but will only be inclined to do so if it is trained on a task demanding this in the first instance. This explains why the PrediNet is able to perform reasonably well on the RM\textrightarrow RP curriculum, but performs poorly on BC\textrightarrow RM.

\textbf{Lacking spatial inductive bias:} A downside of using Slot Attention rather than CNNs is that, whilst the latent representation disentangles x and y position, models will not learn relations which are translation or rotation invariant unless trained to do so. It is primarily\footnote{The fact that the CNN in the PrediNet is able to learn relational encodings directly also contributes, but based on differences in BC\textrightarrow CP, it is unlikely that this is the leading order cause of differences on BC\textrightarrow RM.} for this reason that the Slot PrediNet, even when provided with ground-truth object representations, performs worse than the PrediNet on the BC\textrightarrow RM curriculum.

\textbf{Lack of a null slot in the PrediNet:} We also observed that the PrediNet performs worse than the RelationNet on the non-relational Sort-of-CLEVR tasks. The simplest explanation for this is that the PrediNet's relation learning mechanism has two steps which can discard salient information about objects; namely, the attention mechanism, and the subtraction-based comparator. As each relation head applies the same set of weights to both of its aggregated input slots, it is not trivial for a single head to implement the identity operation. This could be resolved most easily by adding a null slot\footnote{As there are six slots in our SA model, and six objects in all Sort-of-CLEVR scenes, there were no consistently free slots by default.} to the PrediNet's inputs (similar to \citet{goyalNeuralProductionSystems2021a}).

\section{datasets}
\label{sec:app:datasets} 

\begin{figure}[h]
\centering
\includegraphics[width=0.6\textwidth]{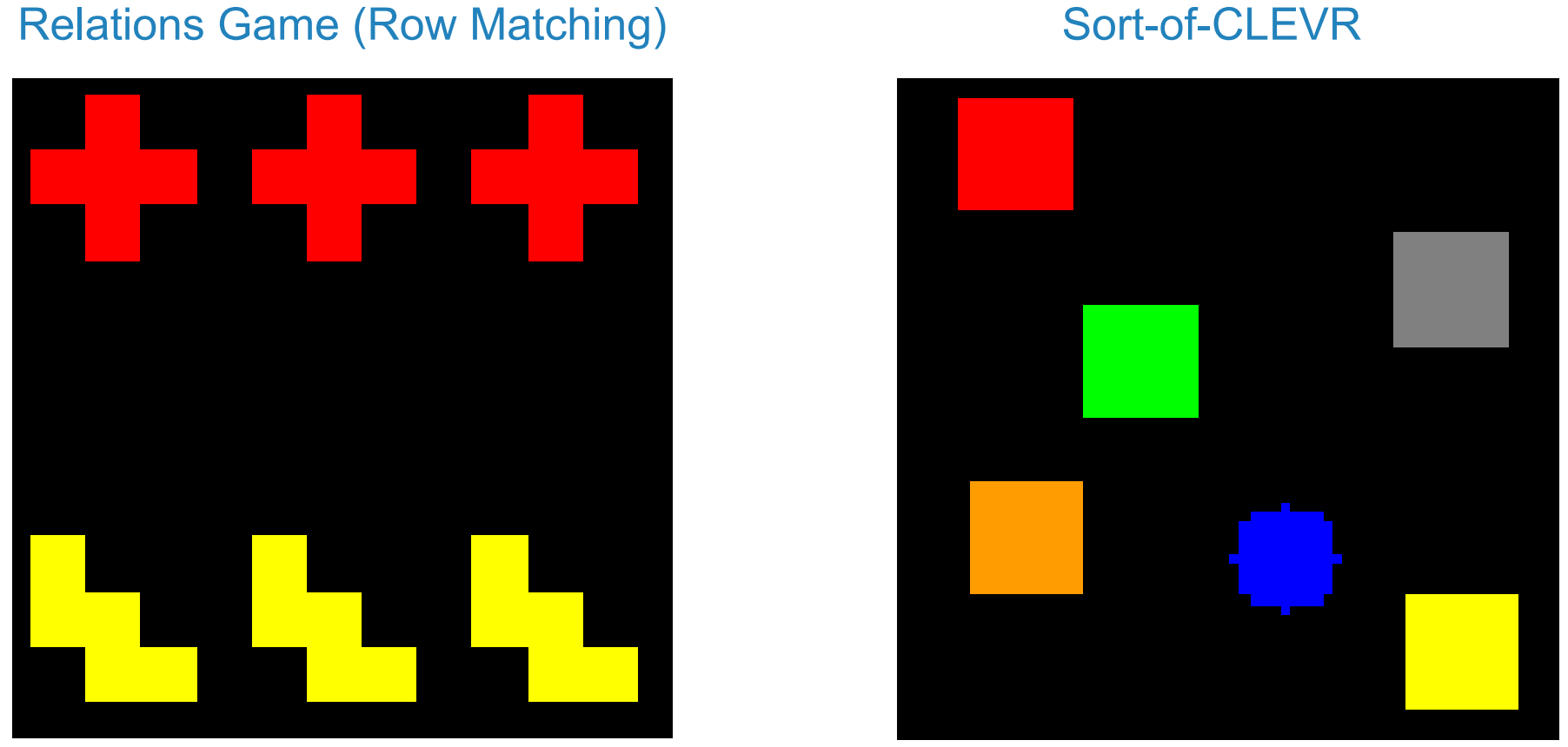}
\caption{Examples of the datasets. On the left, a ``Row Matching'' task from the Relations Game, in which the patterns in both rows (AAA) match, and the label is True. On the right, a Sort-of-CLEVR configuration, which has a set of 20 associated questions. These include non-relational ``What color is the circle"), and relational (``What is the color of the square closest to the circle") questions.}
\label{fig:app:dataset}
\end{figure}

\subsection{Relations Game}
\label{sec:app.datasets.relations_game}

The relations game, proposed by \citet{shanahanExplicitlyRelationalNeural2019}, provides a simple set of relational reasoning problems in a low-dimensional image setting, of which an example is shown in \cref{fig:app:dataset}.

\newpage
As in \citet{shanahanExplicitlyRelationalNeural2019}, we employ multiple curricula which require the rule learning components to learn compositional representations at train time, which are then not altered when generalizing to novel tasks at test time. In particular, the experimental followed is:
\begin{enumerate}
    \item Train the entire architecture (Encoder + Rule Learner + Task Network) on a simple task, such as Between Columns 
    \item Train a new Task Network (with Encoder and Rule Learner frozen) on a more difficult task, such as the Multi-pattern tasks.
\end{enumerate}
In the case of the PrediNet, the CNN encoder is jointly trained in the first stage, whereas for the Slot PrediNet the Encoder is pre-trained (See below). It is important to note that unlike \citet{shanahanExplicitlyRelationalNeural2019} we do not alter the shapes (e.g. from pentominos to hexominos) during the generalisation stage, as we are not interested in assessing encoder generalisation in this work.

\subsubsection{Pretraining Slot Attention}
To pretrain Slot Attention we used the same autoenconding setup as \citet{kipfContrastiveLearningStructured2020}, performing unsupervised reconstruction with an L2 loss.

As the tetrominoes in the relations game are arranged in a regular grid, we created a pretraining dataset in which the positions of the tetrominoes varied continouously, and their sizes were sampled from three possibilities (see \cref{fig:app.slotatt.size.assign}. This ensured that the mappings of the latent space were amenable to the investigation carried out in \cref{sec:app.slotatt.assignment}.

\subsection{Sort-of-CLEVR}
As the name suggests, Sort-of-CLEVR is similar to the CLEVR \cite{clevr} dataset, except that it consists only of 2D circles and squares, with images being of size $75\times75$ and containing six objects. The tasks in Sort-of-CLEVR may be divided into relational and non-relational questions, with relational questions referring to properties of objects in a variety of ways. An example image is shown in \cref{fig:app:dataset}. Our implementation is based on that of \citet{socdataset}, with some modifications to the question encoding format, and the removal of the white background.
\begin{table*}[h]
    \centering
    \caption{Architectural Details for the architectures used. The spatial-broadcast decoder architecture used for slot-attention pre-training was identical to that of \citet{locatelloObjectCentricLearningSlot2020}.} 
    \label{tab:architectural_details}
    \vskip 0.15in
    \begin{center}
    \begin{small}
    \begin{tabular}{lll}
    \dtoprule
    Model & Component(s) & Architectural Details \\ \cmidrule(r){1-1} \cmidrule(lr){2-2} \cmidrule(l){3-3} 
    \multirow{3}{*}{Enconder} & Conv2D & $32$ Channels, $(12,12)$ Kernel Size, $(6,6)$ Stride, $(1,1)$ Padding, ReLU \\
            & Learned Position Embedding & Linear Layers: $[32]$ \\
            & Linear Layers & LayerNorm, Linear Layers: $[32, \text{ReLU}, 32]$ \vspace{0.2cm} \\ 
    
    \multirow{4}{*}{Slot Attention} & Slots &  6 \\
                                    & Latent Sizes & 32 \quad (K,Q,V Projections do not use bias) \\  
                                    & Latent Resolution & Relations Game $(5,5)$; Sort-of-CLEVR $(12,12)$ \\
                                    & Attention Iterations & 3 at train and test \vspace{0.2cm} \\  
                                    
    \multirow{2}{*}{PrediNet}    & Number of Heads & 16 \\
                                 & Relations per Head & 8  \vspace{0.2cm}\\
                
    \multirow{2}{*}{RelationNet} & $g_{mlp}$ & Linear Layers: $[2000,\; \text{ReLU}, \: 2000,\; \text{ReLU}, \: 2000, \;\text{ReLU}, \: 2000,\; \text{ReLU}]$ \\
                             & $f_{mlp}$ & Linear Layers: $[2000,\; \text{ReLU},\: 1000, \;\text{ReLU}, \: 500,\; \text{ReLU},\: 100, \text{ReLU}]$ \vspace{0.2cm}\\
                
    \multirow{2}{*}{Task MLP}  & Relations Game &  Linear Layers: $[8,\; \text{ReLU}, \;2, \;\text{Softmax}]$ \\ 
                               & Sort-of-CLEVR &  Linear Layers: $[16,\; \text{ReLU}, \;10, \;\text{Softmax}]$ \\ \dbottomrule
    \end{tabular}
    \end{small}
    \end{center}
    \vskip -0.1in
\end{table*}

\newpage
\section{Implementation Details}


Architectural details are provided in \cref{tab:architectural_details}. When ground truth inputs are provided to the Slot PrediNet and RelationNet, these are constructed into vectors by:
\begin{enumerate}
    \item Mapping all object properties to the interval $[-1, \:1]$
    \item Concatenating (in some fixed order) these properties for a given object, to form a slot
    \item Stacking the set of slots for all objects present in the scene
    \item Appending empty slots (zeros), so that there are six slots in total (never needed for Sort-of-CLEVR)
\end{enumerate}
Additionally, a learned linear layer with 34 units is applied across these slots, and considered part of the ``Encoder".

\subsection{Libraries}
Decision Trees were implemented using Scikit-learn \cite{scikit-learn}. For neural network training and experimentation we used Jax \cite{jax2018github}, Haiku \cite{haiku2020github}, Optax \cite{optax2020github} and Tensorflow \cite{tensorflow2015-whitepaper}. Decision trees were pruned using Minimal Cost-Complexity~\cite{breimanClassificationRegressionTrees2017}.

\end{document}